\documentclass[letterpaper, 10 pt, journal, twoside]{IEEEtran}

\usepackage{epsfig}
\usepackage{amssymb}
\usepackage{cite}
\usepackage{booktabs}
\usepackage{multirow}
\usepackage{mathtools}
\usepackage{bm}
\usepackage{algorithm}
\usepackage{comment}
\usepackage{color}
\usepackage{setspace}

\definecolor{green}{rgb}{0.01, 0.5, 0.01}

\usepackage{subcaption}
\usepackage{url}
\usepackage{threeparttable}

\usepackage{amsmath}
\usepackage[noend]{algorithmic}
\usepackage{array}

\begin{document}

\begin{titlepage}
\quad\\[1cm]
\makeatother
	{\Huge IEEE Copyright Notice}\\[0.5cm]
	{\begin{spacing}{1.2}
	\large \copyright \ 2022 IEEE. Personal use of this material is permitted. Permission from IEEE must be obtained for all other uses, in any current or future media, including reprinting/republishing this material for advertising or promotional purposes, creating new collective works, for resale or redistribution to servers or lists, or reuse of any copyrighted component of this work in other works.
	\end{spacing}}
\end{titlepage}

\title{Category-Association Based Similarity Matching for \\Novel Object Pick-and-Place Task}
\author{Hao Chen, Takuya Kiyokawa, Weiwei Wan, and Kensuke Harada\vspace{-5pt}
\thanks{Manuscript received: September 9, 2021; Revised: December 7, 2021; Accepted: January 4, 2022.}
\thanks{This paper was recommended for publication by Editor Hong Liu upon evaluation of the Associate Editor and Reviewers. This work was supported by New Energy and Industrial Technology Development Organization (NEDO).}
\thanks{All authors are with the Department of Systems Innovation, Graduate School of Engineering Science, Osaka University, Toyonaka, Osaka, Japan.
        {\tt\small \{h.chen, kiyokawa\}}@hlab.sys.es.osaka-u.ac.jp and {\tt\small\{wan, harada\}}@sys.es.osaka-u.ac.jp}%
\thanks{Digital Object Identifier (DOI): see top of this page.}
}

\markboth{IEEE ROBOTICS AND AUTOMATION LETTERS. PREPRINT VERSION. January, 2022}
{Chen \MakeLowercase{\textit{et al.}}: Similarity Matching for Pick-and-Place} 

\maketitle

\vspace{-20pt}
\begin{abstract}
Robotic pick-and-place has been researched for a long time to cope with uncertainty of novel objects and changeable environments. Past works mainly focus on learning-based methods to achieve high precision. However, they have difficulty being generalized for the limitation of specified training models. To break through this drawback of learning-based approaches, we introduce a new perspective of similarity matching between novel objects and a known database based on category-association to achieve pick-and-place tasks with high accuracy and stabilization. We calculate the category name similarity using word embedding to quantify the semantic similarity between the categories of known models and the target real-world objects. With a similar model identified by a similarity prediction function, we preplan a series of robust grasps and imitate them to plan new grasps on the real-world target object. We also propose a distance-based method to infer the in-hand posture of objects and adjust small rotations to achieve stable placements under uncertainty. Through a real-world robotic pick-and-place experiment with a dozen of in-category and out-of-category novel objects, our method achieved an average success rate of 90.6\% and 75.9\% respectively, validating the capacity of generalization to diverse objects. 
\end{abstract}


\IEEEpeerreviewmaketitle

\section{Introduction}
\IEEEPARstart{A} {\lowercase{c}}omplete vision-based robotic manipulation system contains the modules of object detection, grasp selection and motion planning. Although previous work has studied extensively on each of the modules \cite{Du2021VisionbasedRG}, the accumulated error occurring in sensing, planning and controlling processes has made the whole system difficult to achieve high precision. A few studies have focused on using deep learning to make a highly precise pick-and-place system come true \cite{Gualtieri2018PickAP,Berscheid2020SelfSupervisedLF,KalashnikovIPIH18,Manuelli2019}. However, such of learning-based methods are inevitably costly and time-consuming for training. They also have a critical limitation of precision decrease when the conditions of manipulation tasks change, like using different types of cameras or robots.

To break through such limitations, we propose a training-free method using similarity matching between a known database and unknown targets. Given a novel object under uncertainty, we use only one depth sensor equipped on robot hand to obtain its semantic information and point clouds. Meanwhile, we construct a grasp database which also includes semantic information and point clouds of a few known models. By developing a similarity prediction method, we can identify the specific model with the highest similarity to the novel object from the database. Assumed similar objects share similar robust graspable points, we can first plan a series of robust grasps on a known model, and then transfer them to the real-world target object using point cloud registration. In this way, we can avoid the difficulty of grasp planning in partially-observed situations. Moreover, by applying point cloud registration with known models, reorientation of objects in pick-and-place task is easy to realize as we have no need of estimating the initial posture of the target object at the beginning.

\begin{figure*}[t]
    \centering
    \includegraphics[width=\linewidth]{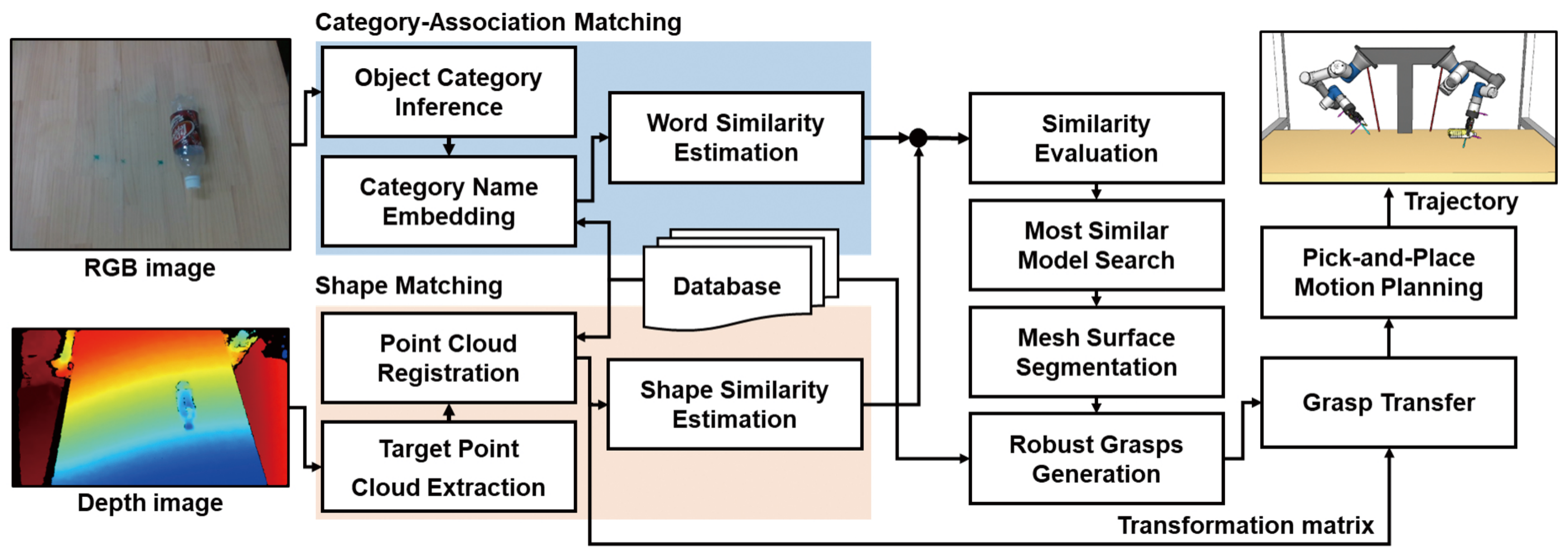} 
    \caption{Overview of the proposed pick-and-place system. The inputs include one RGB image, one depth image and one existing database. The output is a complete pick-and-place motion trajectory executable by the robot.}
    \label{img:1}
\end{figure*}

Our work mainly makes three contributions.

\textbf{Proposal of category-association based similarity prediction combined with point cloud registration.} Recent technologies like YOLOv5 \cite{Redmon2016YouOL} and DeepLabV3+ \cite{Chen2018} have enabled us to extract semantic information from a single image, making it easy to obtain the category name of a novel object with an RGB camera. By introducing a database consisting of the category information of a few known models, we can evaluate the semantic similarity between the categories of novel objects and known models. In addition, we incorporate point cloud registration to evaluate their shape similarity. By combining the two similarity matching results, we can make a reliable prediction of which known model is most similar to the real-world target object.

\textbf{Construction of a novel object pick-and-place system based on imitating grasps.} We construct a vision-based robotic manipulation system aimed at novel objects using the similarity matching results. Each model in the database is preplanned with a series of robust grasps. We use a robotic gripper that is two-finger parallel type in our system and thus the grasps are all antipodal. To achieve pick-and-place tasks with the parallel-jaw gripper, we imitate the preplanned antipodal grasps on known models to plan robust grasps on target objects, and incorporate the idea of rotational adjustment to realize stable placements. Finally we apply the DD-RRT algorithm \cite{Kuffner2000RRTconnectAE} to plan a collision-free trajectory for the motions of the robot.

\textbf{High success rate of novel object pick-and-place based on training-free methods.} Among the previous studies on novel object manipulation, few of them have achieved a very high success rate even combined with deep reinforcement learning \cite{Berscheid2020SelfSupervisedLF}. But through our method, we have demonstrated by experiments that the idea of similarity matching can lead to an average success rate over 90\% with in-category novel objects and around 75\% with out-of-category novel objects. It has outperformed many existing learning-based methods, not to say other training-free methods.

\section{Related works}
Recent approaches of novel object manipulation are mainly divided into two types: image-processing-based and deep-learning-based. Both of them are studied extensively. Another perspective is to make use of prior knowledge and learn a new grasp by imitating existing robust grasps, which is close but different from our proposed method.

\textbf{Image-processing-based novel object pick-and-place.}
Image processing is an important technique in robotic manipulation. Vohra et al. \cite{Vohra2021EdgeAC} propose a novel edge and corner detection algorithm for an unorganized point cloud. They extract pose information from the detection result and create a pick-and-place motion plan based on it. Gualtieri~et~al.~\cite{Gualtieri2021RoboticPW} apply object segmentation and shape completion to refine the point cloud of target objects and use a three-level planner to realize regrasp manipulation \cite{Wan2017RegraspPU}. Such of methods focus on the improvement of perception modules, thus the planning accuracy is limited to the precision of visual processing. 

\textbf{Deep-learning-based novel object pick-and-place.}
Recent studies mainly focus on learning a model to generate manipulation tactics. Gualtier et al. \cite{Gualtieri2018PickAP} incorporate deep reinforcement learning by encoding grasp images as input to evaluate success metrics. Kalashnikov et al. \cite{Berscheid2020SelfSupervisedLF} and Berscheid et al. \cite{KalashnikovIPIH18} develop a self-supervised learning method without human interaction. Manuelli et al. \cite{Manuelli2019} propose a keypoint detection model to realize category-level pick-and-place. Zeng et al. \cite{Zeng2018RoboticPO} develop a pick-and-place system of novel objects in clutter with a cross-domain image matching strategy that does not need additional task-specific training. Gualtieri et al. \cite{Gualtieri2018Learning6G} develop a specific set of constraints for the robot to learn a sequence of gazes to focus attention on the task-relevant parts of the scene. Such learning-based methods are superior in achieving a high success rate in the case that the target object is within the categories of the training dataset. However, when the types of objects vary or the conditions of manipulation tasks change, the success rate will make a remarkable drop. Moreover, they require costly and time-consuming work.

\textbf{Imitation-based grasp planning of novel objects.}
Only a few previous studies focus on imitation approaches to realize novel object manipulation. The Columbia Grasp Database developed by Goldfeder et al. \cite{Goldfeder2009} use Zernike descriptors \cite{Novotni20033DZD} to evaluate the similarity between two mesh models, and share the same grasps with two similar models. Dang et al. \cite{Dang2013GraspAO} use tactile experience generated from a list of shape primitives to adjust grasps to novel objects with similar geometries. Dixit~\cite{Dixit2016AdvancingRG} incorporates the idea of transfer learning to imitate human grasps with a humanoid robotic gripper. The problems of these existing methods are common: they often have extra need for a special type of hand and preparatory work such as tactile sensing before planning grasps, and they are subject to error when the poses of target objects vary.

\begin{figure*}[t] 
    \centering
    \includegraphics[width=\linewidth]{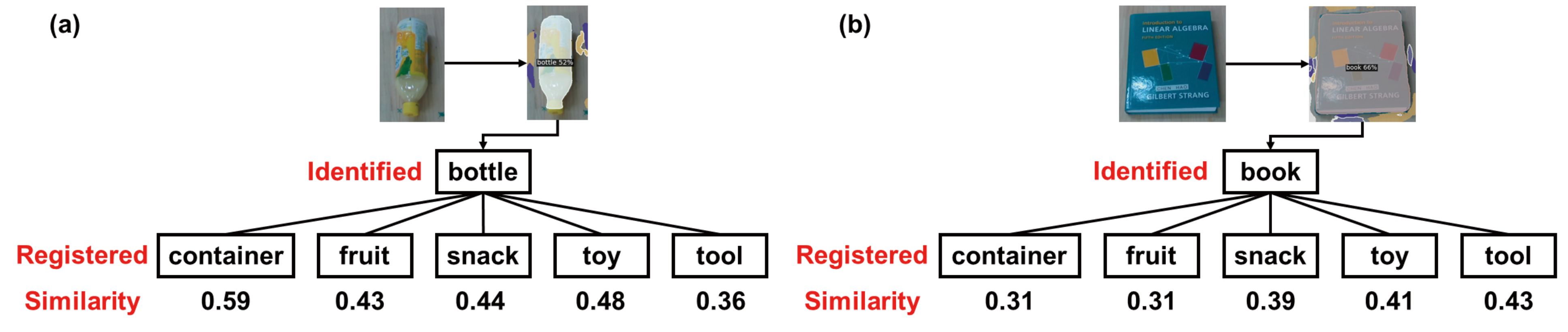} 
    \caption{Example results of CAM using Panoptic FPN and Word2Vec as the approaches of image segmentation and word embedding. In two tests, (a) \textit{bottle} and (b) \textit{book} are detected. Scores shown below represent the category-association matching results between the novel object and the registered models. The word similarity value is calculated in the embedded space.}
    \label{img:2}
\end{figure*}

\section{Similarity Prediction}
\subsection{System Overview}
The overview of our system is shown in Fig. \ref{img:1}. The inputs are one RGB image and one depth image including the novel object. Meanwhile there is an existing database consisting of a few known models with the information of categories, point clouds and preplanned robust grasps. We develop the following method to evaluate the similarity between the known models and target objects.

With the RGB image, we apply object segmentation and word embedding as the techniques of category-association matching to get an evaluated score of semantic similarity. With the depth image, we use PCL (Point Cloud Library)~\cite{Rusu_ICRA2011_PCL} to extract interested point clouds and then use SAC-IA (Sample Consensus Initial Alignment) \cite{Rusu2009FastPF} with ICP (Iterative Closest Point) \cite{Low2004LinearLO} as the approach of point cloud registration to get another evaluated score of shape similarity. Combining the two scores, we can identify the specific model with the highest similarity to the target object from the existing database. With this model, we apply a surface segmentation method \cite{Wan2021PlanningGW} to preplan a series of robust grasps on it. With these grasps, we use the transformation matrix obtained from point cloud registration to transfer them from the model to the target object, and develop an optimization algorithm to minimize the transfer error. Meanwhile, we develop a method of rotational adjustment to infer the in-hand posture of the target object and optimize the placement. Finally we apply the DD-RRT algorithm \cite{Yershova2005DynamicDomainRE} to plan a collision-free trajectory for the pick-and-place motion. In the case that the planning fails in solving IK (Inverse Kinematics), the system will automatically select another grasp from the preplanned grasps and repeat the planning process until success.

\subsection{Category-Association Matching (CAM)}\label{CAM}
Every object has a category which it belongs to, like \textit{apple} belongs to \textit{Fruit}, \textit{cat} belongs to \textit{Animal}. Category is an index used to classify things with similar features, it can give us the association of two objects. For instance, when the word \textit{bottle} is given, we will easily associate it with a word like \textit{water}, but will not relate to a word like \textit{book}. In this case, we consider the two words \textit{bottle} and \textit{water} as high similarity, \textit{bottle} and \textit{book} as low similarity. With this definition, we can give a similarity prediction of category-association between two objects.

In our method, we first apply an object segmentation technique called Panoptic FPN (Feature Pyramid Networks)~\cite{Kirillov2019} derived from Detectron2 library \cite{wu2019detectron2} to extract semantic information from an RGB image. From the extraction result we can identify the category of the target object and then match it with the registered categories in the database using a word embedding technique called Word2Vec \cite{Mikolov2013EfficientEO}. The algorithm of Word2Vec will output a score between 0 and 1 representing the degree of category-association between the target object and registered models. We test several objects to obtain the scores, including in-category objects and out-of-category objects. Example results are shown in Fig. \ref{img:2}. The in-category object, \textit{bottle}, shows a high score of 0.59 with category \textit{container}, low scores with other categories. The out-of-category object, \textit{book}, shows low scores with all registered categories. Other test objects show similar results depending on whether they are in-category or out-of-category.

Two conclusions can be derived from the test. First is, if the novel object belongs to one of the registered categories, we can easily distinguish it from the other categories by the results of CAM. Second is, if the novel object does not belong to any registered category, the results of CAM are not remarkable to classify the object and thus we need another matching method for further similarity evaluation.

\subsection{Point Cloud Registration (PCR)}\label{PCR}
It is obvious that CAM can only help us identify the category of the target object, but is not sufficient to find a precise model with the highest similarity. As objects can be much different even if they belong to the same category, like \textit{bottle} has the type of round-bottom and square-bottom, we still need a method to evaluate the shape similarity. Thus we incorporate PCR in our method.

Before implementing PCR, we have to extract the point cloud of the target object from the complex background. In our case, the target object is in a single state and placed on a fixed desk. The initial position of camera is unchanged before every motion. Based on these assumptions, we can simply use the algorithms of plane segmentation and outlier filter from PCL and set the threshold of coordinates to extract our interested point cloud. In this way, the point cloud of the target object is easy to obtain.

\begin{figure}[t] 
    \centering
    \includegraphics[width=0.98\linewidth]{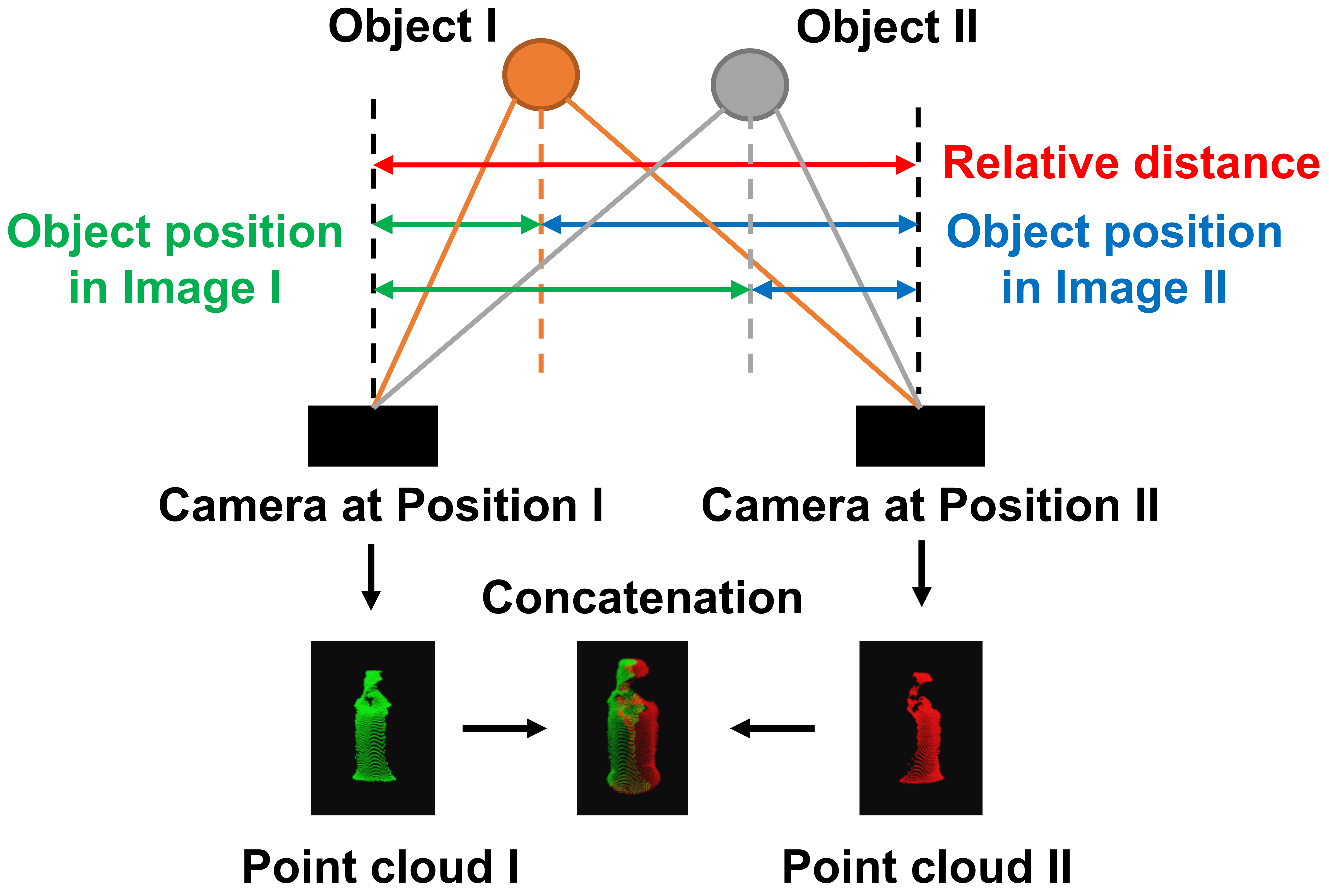} 
    \caption{Invariant relative distance between the object's positions shown in two images when we make the orientation of the camera unchanged. With a certain translation matrix, we can easily concatenate two single point clouds to obtain a more complete point cloud of the target object.}
    \label{img:3} 
\end{figure}

However, another important problem in processing the point cloud is that we can only get a partial view of the target object with only one camera. The incomplete point cloud may lead to unreliability of registration results. To solve this problem, we obtain two point clouds of the target object from different viewpoints by moving the in-hand camera. In this process, we only change the position of camera, but do not change its orientation. As shown in the upper part of Fig. \ref{img:3}, when the camera is set with the same orientation at two positions, wherever the object is, the relative distance of its position shown in two images is invariant. Based on this theory, we can concatenate the two point clouds obtained from different positions using a simple translation matrix. An example of concatenation is shown in the lower part of Fig. \ref{img:3}. Obviously the concatenated point cloud of a bottle is more complete than the single one, thus making the results of PCR more reliable.

\begin{figure}[t] 
    \centering
    \includegraphics[width=0.9\linewidth]{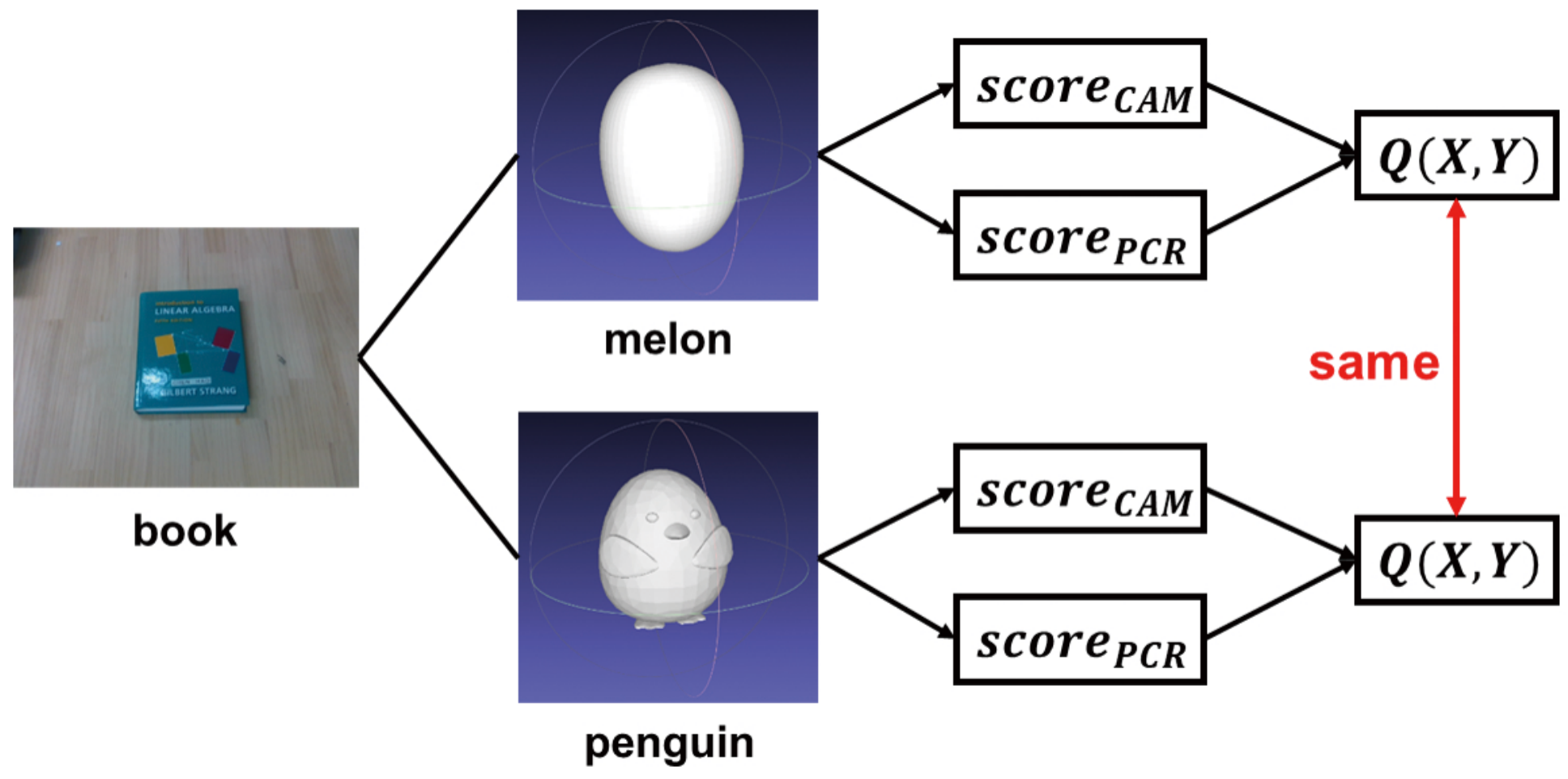} 
    \caption{Identification of the constant coefficients $\mu_{c}$ and $\mu_{p}$. We make use of an out-of-category object and two registered models with similar shapes but belong to different categories to calculate their ${Q}$ values. In this case the two results are assumed as the same.}
    \label{img:4} 
\end{figure}

The algorithm of PCR we use is SAC-IA \cite{Rusu2009FastPF} combined with ICP \cite{Low2004LinearLO}. SAC-IA acts as the role of coarse registration while ICP acts as the role of fine registration. Like CAM, PCR also output a score representing the shape similarity. However, the difference is this score represents the mean of squared distances between two point clouds, which means good fitness will inversely lead to a low score.

\subsection{Similarity Quantification}
Both CAM and PCR are indispensable in our method of similarity prediction. Only CAM is not sufficient to find a precise model. Only PCR is not reliable enough due to the partially-observed situations. With the two scores obtained from the two matchings, we quantify the similarity with a simple function shown as below:
\begin{equation}
    \label{eqn:1}
    Q(X,Y)=\mu_{c}\cdot C(X,Y)+\mu_{p}\cdot P(X,Y),
\end{equation}
where $Q(X,Y)$ is the evaluated similarity between the novel object $X$ and the known model $Y$. $C(X,Y)$ represents the semantic similarity between $X$ and $Y$ related to the result of CAM. $P(X,Y)$ represents the shape similarity between $X$ and $Y$ related to the result of PCR. $\mu_{c}$ and $\mu_{p}$ are constant coefficients that determine the weights of semantic similarity and shape similarity.

As mentioned in Section \ref{CAM}, the score of CAM is a value between 0 and 1. If we directly use it as $C(X,Y)$, we also have to normalize the score of PCR to a value between 0 and 1 to be used as $P(X,Y)$. Through trials we found the scores of PCR are values within the range of 0.0001 to 0.1. Considering the equal distribution between 0 and 1, we introduce logarithm in the calculation of $P(X,Y)$. Also considering low score represents high similarity, we identify the two parameters with the following equations:
\begin{eqnarray}
\label{eqn:2}
C(X,Y)&=&{score}_{CAM}, \\
P(X,Y)&=&\frac{{\rm log}\;0.1-{\rm log}\;{score}_{PCR}}{{\rm log}\;0.1-{\rm log}\;0.0001}.
\end{eqnarray}

Another essential problem is the setting of $\mu_{c}$ and $\mu_{p}$. In Section \ref{CAM}, we have tested that the out-of-category object \textit{book} will output similar scores of CAM when matching with the registered categories in the database, which means \textit{book} has similar semantic similarity with all the registered categories. In this case, if we find two models in different categories but with similar shapes from the database, they are supposed to have both similar semantic similarity and shape similarity with the object \textit{book}, thus the final calculated ${Q}$ values should be almost the same. Based on this theory, we select the model \textit{melon} from the category \textit{fruit} and the model \textit{penguin} from the category \textit{toy} to do similarity matching with the object \textit{book}, as shown in Fig. \ref{img:4}, and identify the constant coefficients $\mu_{c}$ and $\mu_{p}$ with the following equations:
\begin{eqnarray}
&Q(melon,book) = Q(penguin,book), \\
&\mu_{c}+\mu_{p}=1.
\end{eqnarray}

The results can be easily obtained as $\mu_{c}=0.52, \mu_{p}=0.48$. We input the two values into Equation (\ref{eqn:1}) and test the results with some other out-of-category objects (e.g. \textit{toothbrush}). Similarly we match them with two registered models with similar shapes but belong to different categories (e.g. \textit{orange} in the category \textit{fruit} and \textit{ball} in the category \textit{toy}). The calculated $Q$ values still appear to be very close, which confirms the reasonability of obtained constant values. Then we can find the most similar model by ranking all the $Q(X,Y)$:
\begin{equation}
    \label{eqn:4}
    Y^*={\rm argmax}(Q(X,Y)),
\end{equation}
where $Y^*$ represents the specific model obtaining the highest value of $Q(X,Y)$.

\section{Pick-And-Place Planning}
\subsection{Grasp Transfer}
During the process of PCR, besides the score of similarity we also get a transformation matrix between the model and the real-world object. With this matrix we can transfer the preplanned robust grasps on known models to the similar grasps on novel objects. Thus we first apply an approach of grasp planning based on the superimposed segmentation of object meshes \cite{Wan2021PlanningGW} to preplan a series of robust grasps, and then apply the transformation matrix from PCR to transfer them from registered models to real-world objects to realize imitation learning of robust grasps. 

\begin{figure}[t] 
    \centering
    \includegraphics[width=0.9\linewidth]{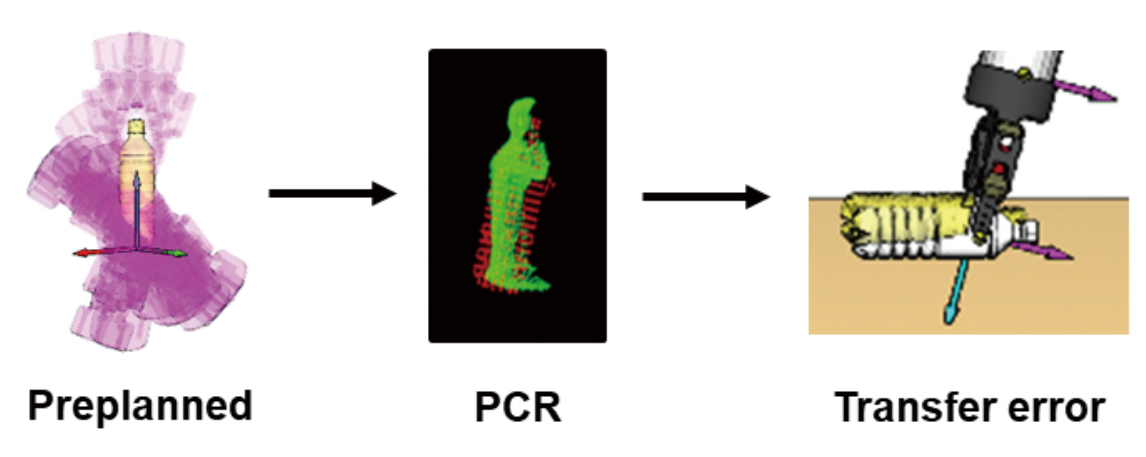} 
    \caption{Preplanned grasps on a known model and error in PCR and grasp transfer. (Left) A model is preplanned with a series of robust grasps. (Middle) In PCR, the point cloud of the registered model (red) is not perfectly aligned with the partially-observed point cloud of the target object (green). (Right) Due to the error of the transformation matrix, the calculated position of the target object (yellow) deviates from the real position (white).}
    \label{img:5}
\end{figure}

However, the real condition is that the transferred grasps are subject to error due to the inaccuracy of PCR. As shown in Fig. \ref{img:5}, the incomplete point cloud of the target object (green) is not perfectly aligned with the point cloud of its similar model (red) in position and orientation due to the partial observation. Thus the transformation matrix obtained from PCR is inevitably subject to error. With this imprecise matrix, the calculated position of the novel object (yellow) deviates from its real position (white). In this case the transferred grasp has a possibility of failing if the error is large.

To reduce the error, we first obtain point clouds of the target object from several viewpoints and apply PCR to get a set of transformation matrices. We extract the rotation and translation information from the matrices and calculate the mean results of position and orientation of the grasps. Among them we remove results that obviously deviate from other results (orientation error $>$ 90°) and obtain the final transformation matrix $T$.

To improve the precision further, we develop an optimization algorithm to minimize the error. With the matrix $T$, we compare its error to a threshold $\delta$. If the error is small enough, the algorithm ends, otherwise we move the camera to a new viewpoint to do calculation again and update the final transformation matrix to be $T^{'}$. Then we check the error of the new matrix. If the error increases, we discard this result and move the camera to another new viewpoint. If the error decreases, we compare it to the threshold $\delta$ again. If the error is small enough, the algorithm ends, otherwise we reserve this result and move the camera to another new viewpoint. We repeat this process until the error becomes small enough.

In our method, we use the score of PCR to represent transfer error. Low PCR scores mean high fitness of two point clouds, which also means the error occurring in the process of grasp transfer is small. The threshold $\delta$ is set to be 0.0001 (same as the lower bound of PCR scores). The optimization algorithm ends when the newly obtained score is lower than the threshold.

\begin{figure*}[t] 
    \centering
    \includegraphics[width=0.9\linewidth]{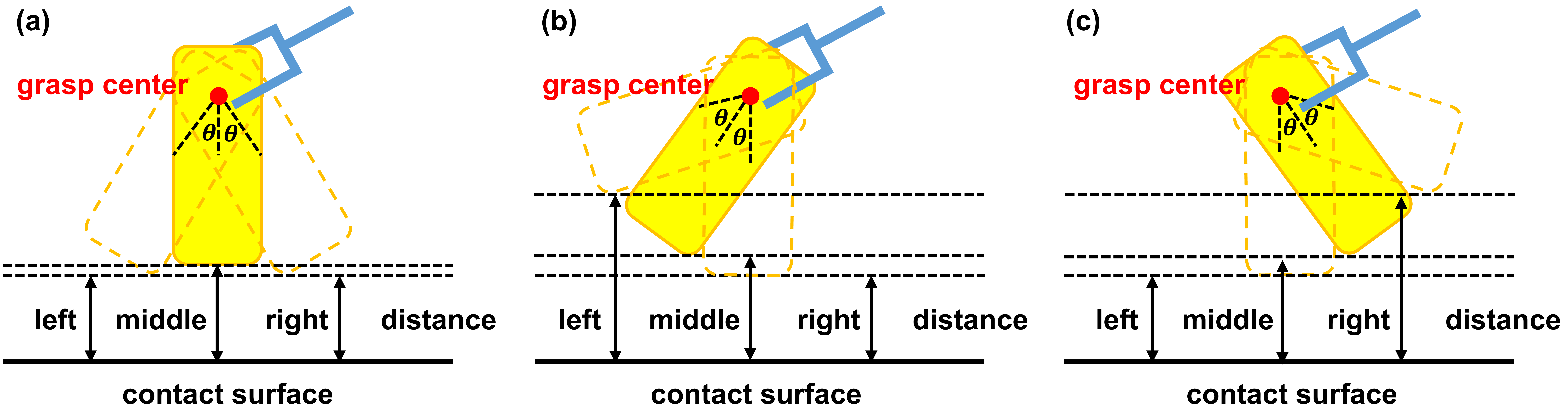} 
    \caption{Inference of the in-hand posture. We rotate the in-hand object with a small angle $\theta$ in two inverse directions by the grasp center. When the original posture is upright (a), the distance between the lowest point of the object and the contact surface does not differ much before and after rotations. When the original posture is right-leaning (b) or left-leaning (c), the distance will differ much in one of the rotations.}
    \label{img:6} 
\end{figure*}

\subsection{Stable Placement}\label{SP}
Above we have solved the problem of grasping, but a remaining problem is how to place the object stably in desired posture. Unlike general methods, we do not need an extra method to estimate the initial posture of object like \cite{Newbury2021LearningTP}. Taking the advantage of a known database, we can simply set the final posture of the target object according to its similar model. However, due to the inevitable error in grasp transfer, the final placement has the risk of being unstable as the in-hand posture of the target object may not be so good as in ideal condition. 

Considering an in-hand model for analysis of stable placement shown in Fig. \ref{img:6}, we assume an object with a regular shape like a square. To find a stable posture, we rotate the object with a small angle $\theta$ in two inverse directions by the grasp center. The rotational plane is selected to be parallel with the two-finger gripper because the object's posture is mostly undetermined in this 2D plane. As seen from the figure, when the object is in an upright posture, the lowest point of the object after rotation in two directions almost share the same distance to the contact surface. But when the object is in a right-leaning or left-leaning posture, the distance becomes much different in one of the rotational directions. 

The difference in distance become the key of inferring the in-hand posture of the target object. To measure the distance, we move the object right down to contact the surface in different rotations and obtain the timing of collision by a force sensor. If the time cost before collision is close in different rotations, the in-hand object is supposed to be in an upright posture. If the time cost is much larger in one of the rotations, the in-hand object is supposed to be right-leaning or left-leaning. In our method, we set the rotation angle $\theta$ to be 30°. After inferring the actual in-hand posture, we adjust the object with a suitable rotation $\gamma$ ($\gamma \in \left \{-\theta, 0, \theta \right \}$) to be placed stably.

It should be noted that the proposed method of stable placement is based on the following assumptions:

1) The in-hand posture of the object does not deviate much from the upright posture (tilt angle $<$ 90°);

2) The object hardly slips in horizontal direction when collided with contact surface;

3) The bottom of the object is basically flat.

In our case, the error of grasp transfer is minimized thus the first assumption is satisfied. The second assumption will not be a problem when the rotation angle is not set too large. The third assumption seems to be a limitation to the type of objects. But actually, an object with an uneven bottom is difficult to achieve stable placements, thus not considered in our pick-and-place task.

\subsection{Motion Planning}
After identifying the initial grasp and the final placement, we finally generate a collision-free trajectory of pick-and-place motion using the DD-RRT algorithm \cite{Yershova2005DynamicDomainRE}. However, due to the uncertainty in the posture of preplanned grasps, the motion planning has a possibility of failing in solving IK (i.e. the calculated position and orientation is unreachable by the robot). In this case, we iterate the algorithm with another grasp until the planning succeeds. The framework of our algorithm is shown in Algorithm \ref{alg:1}. To ensure a feasible path can be output, sufficient robust grasps are preplanned for each registered model.
\renewcommand{\algorithmicrequire}{\textbf{Input:}}
\renewcommand{\algorithmicensure}{\textbf{Output:}}
\begin{algorithm}[tb]
\caption{Pick-and-Place Motion Planning}\label{alg:1}
\begin{algorithmic}[1]
\REQUIRE ~~\\
A set of preplanned grasps on the known model, $G$;\\ 
Transformation matrix obtained from PCR, $T$; \\
Final posture of the novel object, ${p_f}$; 
\ENSURE ~~\\
Pick-and-place motion trajectory, $M$ 
\STATE Initial posture of the novel object ${p_i}$ is known by ${T}$
\FOR {each grasp ${g\in{G}}$}
\STATE ${M}\gets {\rm DDRRT}\left \{ g,p_i,p_f \right \}$
\IF {${M}$ is None}
\STATE \textbf{continue}
\ELSE 
\STATE \textbf{break}
\ENDIF 
\ENDFOR
\RETURN ${M}$
\end{algorithmic}
\end{algorithm}

\section{Experiments}
\begin{figure}[t] 
    \centering
    \includegraphics[width=0.6\linewidth]{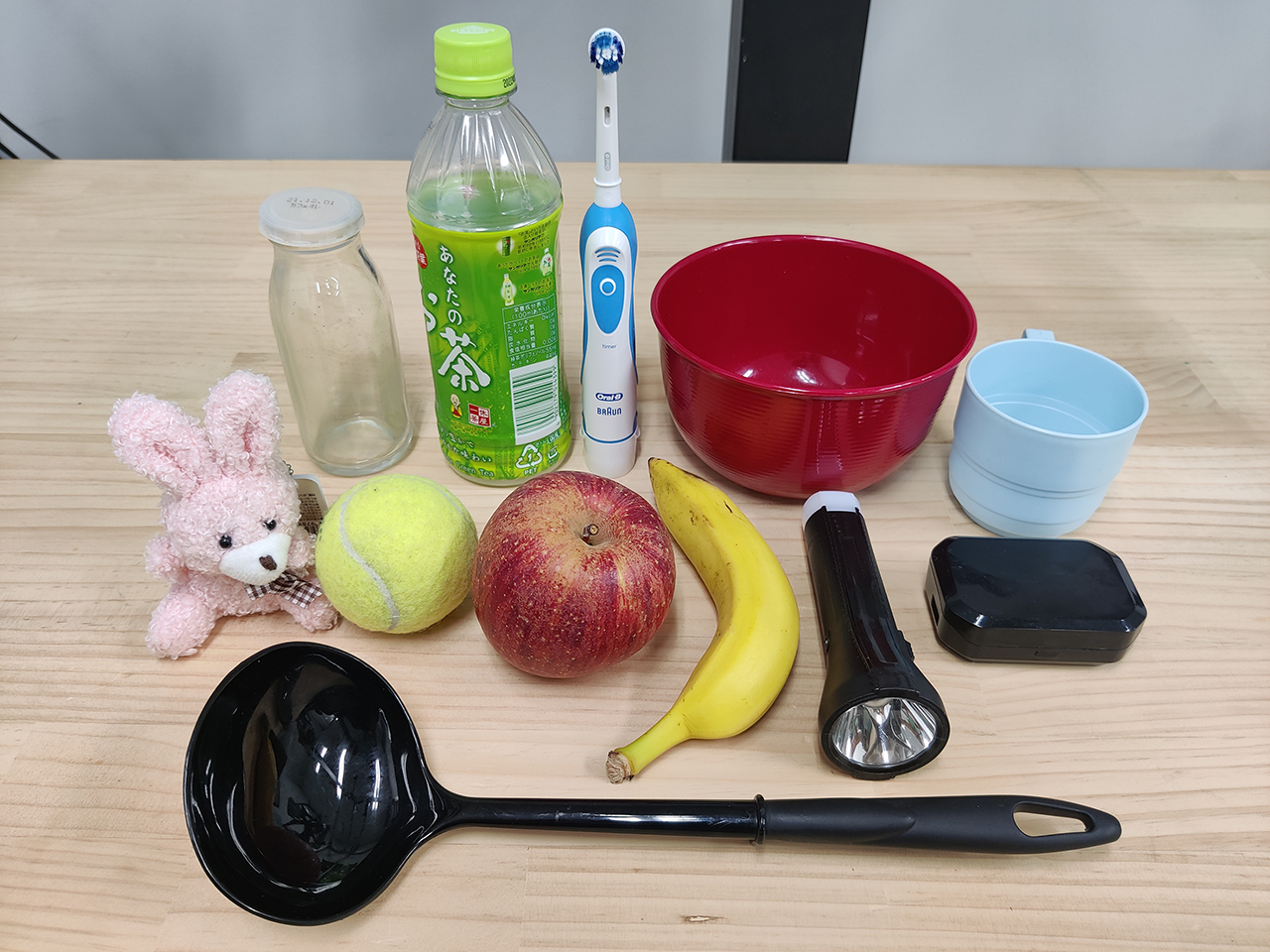} 
    \caption{A dozen of novel objects used for validation experiments. For each object we find an evaluated most similar model from the database and imitate its preplanned robust grasps to achieve pick-and-place tasks.}
    \label{img:7} 
\end{figure}

To verify our method, we perform real-world experiments of robotic pick-and-place with 12 novel objects, as shown in Fig. \ref{img:7}. Among them some are not included in the registered categories (e.g. \textit{toothbrush}), considered as out-of-category novel objects. Others like \textit{apple} included in the category \textit{fruit} are considered as in-category novel objects.


\subsection{Experimental Setup}
The configuration of our experiment is one Realsense D435 depth camera, one single-armed UR3e robot, and one Robotiq 2F-85 gripper. The novel object is posed in arbitrary position and orientation on a fixed desk.

Meanwhile, we construct a database composed of 100 models in five different categories: \textit{container, fruit, snack, toy, tool}. The registered information includes the categories, point clouds and preplanned antipodal grasps of the models.

\begin{figure}[t] 
    \centering
    \includegraphics[width=\linewidth]{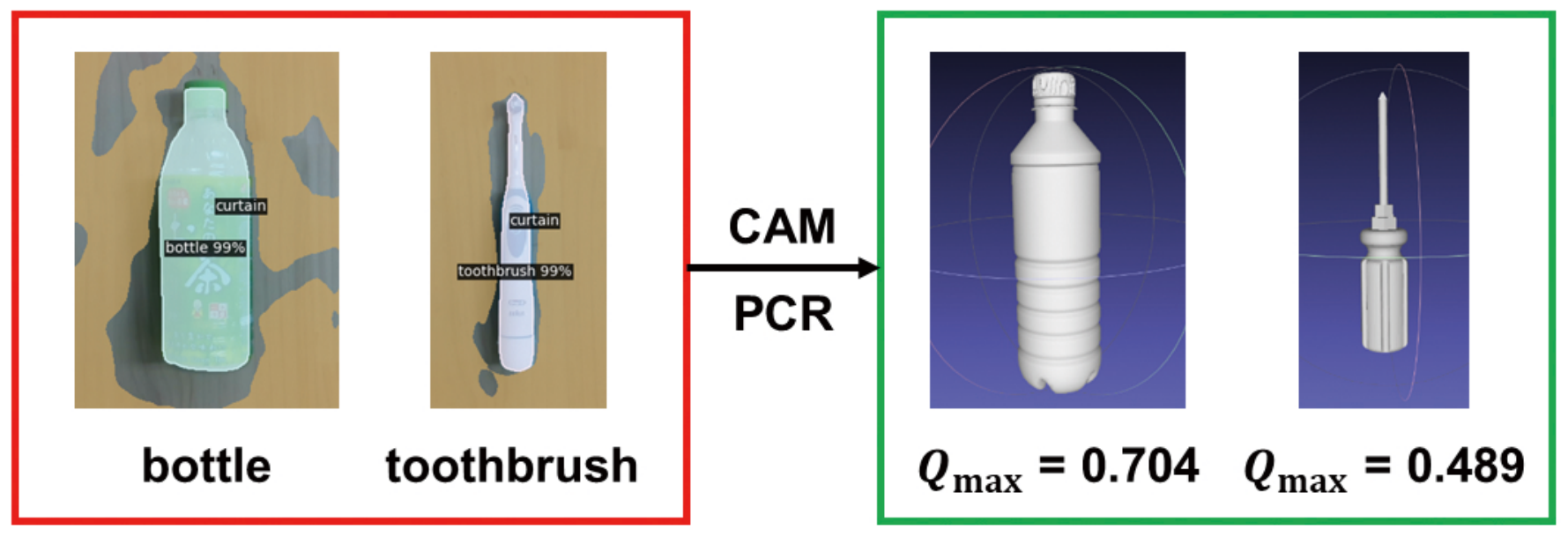} 
    \caption{Example results of similarity prediction based on CAM and PCR. The red block represents the results of semantic identification by Panoptic FPN. The green block shows the most similar models found with the highest \textit{Q} values.}
    \label{img:8}
    \vspace{-5pt}
\end{figure}

\begin{table*}[t]
\small
\renewcommand\arraystretch{1.5}
\setlength\tabcolsep{3pt}
\centering
\caption{Experimental results of novel object pick-and-place.}
\begin{tabular}{r c c c c c c c c | c c c c}
\toprule
                            & \multicolumn{8}{c}{In-category objects} & \multicolumn{4}{c}{Out-of-category objects}                                             \\ 
\hline
Semantic identification & bottle & bottle & cup & bowl & apple & banana & teddy bear & sports ball  & toothbrush & spoon & flashlight & cell phone\\
Total experiments       & 21     & 23    & 20  & 19  & 23   & 20                    & 22    & 22     & 21   & 19     & 23    & 20                     \\
Successful times        & 19     & 20   & 18   & 18   & 19   & 20    & 22     & 18         & 14       & 18     & 19    & 12                  \\ 
\hline
Success rate (\%)       & 90.5   & 87.0  & 90.0 & 94.7  &82.6
& \textbf{100}   & \textbf{100}   & 81.8   & 66.7  
& \textbf{94.7}   & 82.6  & 60  \\
\bottomrule
\end{tabular}
\label{tab:1}
\vspace{-5pt}
\end{table*}

\subsection{Similarity Prediction}
We use the depth camera to obtain RGB and depth images including the novel object. With RGB images, we apply image segmentation to extract the category information of novel objects. An example is shown in the red block of Fig. \ref{img:8}. The in-category object \textit{bottle} and the out-of-category object \textit{toothbrush} are both correctly identified. With the category names obtained, we input them into CAM to obtain the score of semantic similarity between the novel objects and known models in the database. With depth images, we extract the point clouds of novel objects from the complex background by PCL and input them into PCR to obtain the score of shape similarity. Combining the two scores, we calculate the \textit{Q} value by Equation (\ref{eqn:1}) for each matching and find the most similar model with the highest \textit{Q} value. As shown in the green block of Fig. \ref{img:8}, a \textit{bottle} model and a \textit{screwdriver} model are found with the highest similarity to the object \textit{bottle} and \textit{toothbrush} respectively.

In our experiments, the in-category objects all succeed in finding a similar model both in semantic level and geometric level from the database. The out-of-category objects behave differently in similarity prediction. Although Fig. \ref{img:8} shows a positive result of similarity prediction with the out-of-category object \textit{toothbrush}, it is not always the case. Due to the limitation of image segmentation methods and partially-observed situations, the results of CAM and PCR are not always reliable. For instance, the out-of-category object \textit{black box} used in our experiment is wrongly identified as \textit{cell phone}, which is not the truth. And the final matching result comes to be a \textit{candy} model in the category \textit{snack}, which is non-similar (a negative result of similarity prediction). We also experiment on such cases to see to which extent the success rate will differ.

\begin{figure}[t] 
    \centering
    \includegraphics[width=\linewidth]{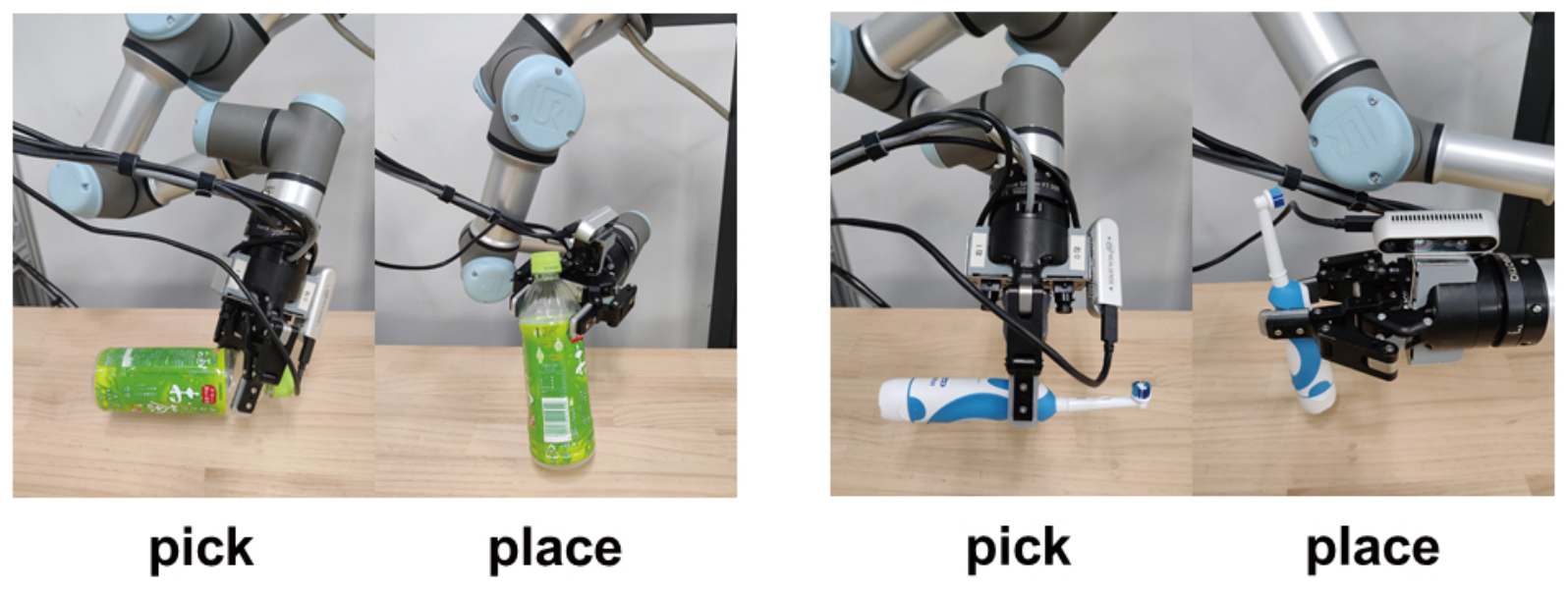} 
    \caption{Realization of pick-and-place motions based on the results of similarity prediction. The grasp postures of the bottle (left) and the toothbrush (right) are generated from the preplanned grasps on the models of a bottle and a screwdriver shown in Fig. \ref{img:8}.}
    \label{img:9}
    \vspace{-5pt}
\end{figure}

\subsection{Novel Object Pick-and-Place}
With the reference models identified by similarity prediction, we preplan over 100 robust grasps on each of them and transfer these grasps from the models to real-world objects with the transformation matrix obtained from PCR. We incorporate the DD-RRT algorithm to plan an IK-solvable collision-free trajectory to reorient the target object on the basis of the transferred grasps, and adjust small rotations to achieve stable placements.

For each object, we experiment about 20 times of pick-and-place motion with arbitrary initial position and orientation. Our goal is to successfully grasp the target and place it at a specified position stably (no movement occurs after placement). The final posture is determined to be the same as the posture of the reference model. Either the failure of grasping or placing will be counted as failure cases. Fig. \ref{img:9} shows two successful cases of pick-and-place with the in-category object \textit{bottle} and the out-of-category object \textit{toothbrush}. From the total experimental results shown in Table \ref{tab:1}, the in-category objects show a high average success rate of 90.6\%, while the out-of-category objects also perform not badly with an average success rate of 75.9\%. 

Some interesting facts are found from the experimental results. The two bottles (one plastic bottle and one glass bottle) both show a high success rate of manipulation in our experiment. As we know, transparent objects are difficult to recognize in depth sensors, thus are difficult to plan robust grasps based on their point clouds. But through our method incorporating semantic identification, they can find similar models in the category \textit{bottle} to imitate their robust grasps, thus performing much better than traditional methods. The \textit{apple} and the \textit{sports ball} both show a relatively low success rate among in-category objects. This is due to their round shapes. Although we have minimized the error in grasp transfer, it is inevitable that the contact points on the real-world objects are not completely consistent with the contact points on the reference models. In this case, slipping is more likely to occur when grasping objects with round shapes.

Moreover, we find the out-of-category objects differ much in success rate of manipulation. The \textit{spoon} achieves the highest success rate due to its similarity matching result to be a \textit{spanner}, which shares similar contact points. The flashlight takes the second place for the same reason. However, the other two objects, \textit{toothbrush} and \textit{black box} perform much worse. The shape of the \textit{toothbrush} determines it to be difficult to place stably in a standing posture. Although it can be grasped successfully in most cases, it still fail many times when executing the placement. The \textit{black box} is wrongly identified as a \textit{cell phone} and its similarity matching result is a \textit{candy}, which is actually non-similar. An interesting phenomenon is even if in this case, the success rate does not become very low. The reason can be analyzed as follows:

When the identified category name does not belong to any registered category in the database, the scores of CAM are often very low. In this case, the final matching result strongly depends on the scores of PCR. As the PCR score represents the mean of squared distances between two point clouds, a model with a smaller size is likely to get lower scores (which will lead to larger \textit{Q} values, referred to Equation (\ref{eqn:2})) when matched with a partially-observed novel object. Thus the reference model found by similarity prediction is likely to be smaller than the real-world object. As PCR will align two point clouds together, the preplanned robust grasps on a smaller model have a high possibility of being placed within the graspable area of a larger object after transfer. In this case, even if the selected model is non-similar to the real-world object, the transferred grasps still have a fairly high success rate of manipulation. This is considered to be a significant advantage of our method in dealing with completely novel objects.

\section{Discussion}
Through experiments we verify our proposed method to be able to achieve a high success rate of novel object manipulation. There are two remarkable advantages of our method compared with other related works. One is the generalization, the techniques used in our system: image segmentation (Panoptic FPN), word embedding (Word2Vec), and point cloud registration (SAC-IA with ICP) are all replaceable. Moreover, our method can deal with both in-category objects and out-of-category objects. Although the similarity prediction are not correct at all time, its precision can be raised by enlarging the database to increase the possibility of a novel object finding a similar registered model. Another is the simplicity, we only use one depth camera and one robot hand to achieve the whole pick-and-place task. Our system is easy to construct and all the hardware can be substituted with other types. 

\section{Conclusion and Future Work}
In this paper, we propose a new perspective of novel object manipulation planning by imitating existing robust grasps. We combine category-association matching with point cloud registration to evaluate the semantic and shape similarity between novel objects and known models. We transfer the preplanned robust grasps from registered models to real-world objects and develop an optimization algorithm to minimize the transfer error. We also incorporate the idea of rotational adjustment to infer the actual posture of in-hand objects and achieve stable placements.

In the future, we will put effort in improving the precision of similarity prediction and grasp transfer. We are considering using more semantic features not only the category but also some other properties such as the composition of objects to strengthen the matching result. We will also explore a new way of shape completion without learning a model to raise the reliability of point cloud registration in our method.

\section{Acknowledgement}
\small{
This research is subsidized by New Energy and Industrial Technology Development Organization (NEDO) under a project JPNP20016. This paper is one of the achievements of joint research with and is jointly owned copyrighted material of ROBOT Industrial Basic Technology Collaborative Innovation Partnership.

We also would like to thank Dr. Yoshiaki Ando and Dr. Floris Erich of AIST for helping creation of the database with a 3D shape measurement system of AIST.}

\bibliographystyle{IEEEtran}
\footnotesize
\bibliography{bibliography/reference}

\begin{thebibliography}{10}
\providecommand{\url}[1]{#1}
\csname url@rmstyle\endcsname
\providecommand{\newblock}{\relax}
\providecommand{\bibinfo}[2]{#2}
\providecommand\BIBentrySTDinterwordspacing{\spaceskip=0pt\relax}
\providecommand\BIBentryALTinterwordstretchfactor{4}
\providecommand\BIBentryALTinterwordspacing{\spaceskip=\fontdimen2\font plus
\BIBentryALTinterwordstretchfactor\fontdimen3\font minus
  \fontdimen4\font\relax}
\providecommand\BIBforeignlanguage[2]{{%
\expandafter\ifx\csname l@#1\endcsname\relax
\typeout{** WARNING: IEEEtran.bst: No hyphenation pattern has been}%
\typeout{** loaded for the language `#1'. Using the pattern for}%
\typeout{** the default language instead.}%
\else
\language=\csname l@#1\endcsname
\fi
#2}}

\bibitem{Du2021VisionbasedRG}
G.~Du, K.~Wang, S.~Lian, and K.~Zhao, ``Vision-based robotic grasping from
  object localization, object pose estimation to grasp estimation for parallel
  grippers: a review,'' \emph{Artificial Intelligence Review (AIR)}, vol.~54,
  2021.

\bibitem{Gualtieri2018PickAP}
M.~Gualtieri, A.~T. Pas, and R.~W. Platt, ``Pick and place without geometric
  object models,'' in \emph{Proceedings of the IEEE International Conference on
  Robotics and Automation (ICRA)}, 2018, pp. 7433--7440.

\bibitem{Berscheid2020SelfSupervisedLF}
L.~Berscheid, P.~Meissner, and T.~Kr{\"o}ger, ``Self-supervised learning for
  precise pick-and-place without object model,'' \emph{IEEE Robotics and
  Automation Letters (RA-L)}, vol.~5, pp. 4828--4835, 2020.

\bibitem{KalashnikovIPIH18}
D.~Kalashnikov, A.~Irpan, P.~Pastor, J.~Ibarz, A.~Herzog, E.~Jang, D.~Quillen,
  E.~Holly, M.~Kalakrishnan, V.~Vanhoucke, and S.~Levine, ``Scalable deep
  reinforcement learning for vision-based robotic manipulation,'' in
  \emph{Proceedings of the 2nd Annual Conference on Robot Learning (CoRL)},
  vol.~87, 2018, pp. 651--673.

\bibitem{Manuelli2019}
L.~Manuelli, W.~Gao, P.~Florence, and R.~Tedrake, ``{kPAM}: Keypoint
  affordances for category-level robotic manipulation,'' in \emph{Proceedings
  of the International Symposium on Robotics Research (ISRR)}, 2019.

\bibitem{Redmon2016YouOL}
J.~Redmon, S.~Divvala, R.~Girshick, and A.~Farhadi, ``You only look once:
  Unified, real-time object detection,'' in \emph{IEEE/CVF Conference on
  Computer Vision and Pattern Recognition (CVPR)}, 2016, pp. 779--788.

\bibitem{Chen2018}
L.-C. Chen, Y.~Zhu, G.~Papandreou, F.~Schroff, and H.~Adam, ``Encoder-decoder
  with atrous separable convolution for semantic image segmentation,'' in
  \emph{Proceedings of the IEEE/CVF Conference on Computer Vision and Pattern
  Recognition (CVPR)}, 2018.

\bibitem{Kuffner2000RRTconnectAE}
J.~Kuffner and S.~LaValle, ``{RRT}-connect: An efficient approach to
  single-query path planning,'' in \emph{IEEE International Conference on
  Robotics and Automation (ICRA)}, vol.~2, 2000, pp. 995--1001.

\bibitem{Vohra2021EdgeAC}
M.~Vohra, R.~Prakash, and L.~Behera, ``Edge and corner detection in unorganized
  point clouds for robotic pick and place applications,'' in \emph{Proceedings
  of the IEEE International Conference on Informatics in Control, Automation
  and Robotics (ICINCO)}, 2021.

\bibitem{Gualtieri2021RoboticPW}
M.~Gualtieri and R.~W. Platt, ``Robotic pick-and-place with uncertain object
  instance segmentation and shape completion,'' \emph{IEEE Robotics and
  Automation Letters (RA-L)}, vol.~6, pp. 1753--1760, 2021.

\bibitem{Wan2017RegraspPU}
W.~Wan and K.~Harada, ``Regrasp planning using 10,000s of grasps,'' in
  \emph{Proceedings of the IEEE/RSJ International Conference on Intelligent
  Robots and Systems (IROS)}, 2017, pp. 1929--1936.

\bibitem{Zeng2018RoboticPO}
A.~Zeng, S.~Song, K.-T. Yu, E.~Donlon, F.~R. Hogan, M.~Bauz{\'a}, D.~Ma,
  O.~Taylor, M.~Liu, E.~Romo, N.~Fazeli, F.~Alet, N.~C. Dafle, R.~Holladay,
  I.~Morona, P.~Q. Nair, D.~Green, I.~Taylor, W.~Liu, T.~A. Funkhouser, and
  A.~Rodriguez, ``Robotic pick-and-place of novel objects in clutter with
  multi-affordance grasping and cross-domain image matching,'' in
  \emph{Proceedings of the IEEE International Conference on Robotics and
  Automation (ICRA)}, 2018, pp. 1--8.

\bibitem{Gualtieri2018Learning6G}
M.~Gualtieri and R.~W. Platt, ``Learning 6-dof grasping and pick-place using
  attention focus,'' in \emph{Proceedings of the 2nd Conference on Robot
  Learning (CoRL)}, 2018, pp. 477--486.

\bibitem{Goldfeder2009}
C.~Goldfeder, M.~Ciocarlie, H.~Dang, and P.~K. Allen, ``The {Columbia} grasp
  database,'' in \emph{Proceedings of the IEEE International Conference on
  Robotics and Automation (ICRA)}, 2009, pp. 1710--1716.

\bibitem{Novotni20033DZD}
M.~Novotni and R.~Klein, ``{3D} {Z}ernike descriptors for content based shape
  retrieval,'' in \emph{Proceedings of the 8th ACM Symposium on Solid Modeling
  and Applications}, 2003, pp. 216–--225.

\bibitem{Dang2013GraspAO}
H.~Dang and P.~Allen, ``Grasp adjustment on novel objects using tactile
  experience from similar local geometry,'' in \emph{Proceedings of the
  IEEE/RSJ International Conference on Intelligent Robots and Systems (IROS)},
  2013, pp. 4007--4012.

\bibitem{Dixit2016AdvancingRG}
S.~Dixit, ``Advancing robotic grasp planning using human heuristics for grasp
  similarity,'' Master's thesis, Oregon State University, 2016.

\bibitem{Rusu_ICRA2011_PCL}
R.~B. Rusu and S.~Cousins, ``{3D} is here: Point {C}loud {L}ibrary ({PCL}),''
  in \emph{Proceedings of the IEEE International Conference on Robotics and
  Automation (ICRA)}, 2011.

\bibitem{Rusu2009FastPF}
R.~Rusu, N.~Blodow, and M.~Beetz, ``Fast {P}oint {F}eature {H}istograms
  ({FPFH}) for {3D} registration,'' in \emph{Proceedings of the IEEE
  International Conference on Robotics and Automation (ICRA)}, 2009, pp.
  3212--3217.

\bibitem{Low2004LinearLO}
K.-L. Low, ``Linear least-squares optimization for point-to-plane {ICP} surface
  registration,'' in \emph{Proceedings of the Technical Report (TR)}, 2004.

\bibitem{Wan2021PlanningGW}
W.~Wan, K.~Harada, and F.~Kanehiro, ``Planning grasps with suction cups and
  parallel grippers using superimposed segmentation of object meshes,''
  \emph{IEEE Transactions on Robotics (T-RO)}, vol.~37, pp. 166--184, 2021.

\bibitem{Yershova2005DynamicDomainRE}
A.~Yershova, L.~Jaillet, T.~Sim{\'e}on, and S.~LaValle, ``Dynamic-{D}omain
  {RRT}s: Efficient exploration by controlling the sampling domain,'' in
  \emph{Proceedings of the IEEE International Conference on Robotics and
  Automation (ICRA)}, 2005, pp. 3856--3861.

\bibitem{Kirillov2019}
A.~Kirillov, K.~He, R.~Girshick, C.~Rother, and P.~Dollar, ``Panoptic
  {S}egmentation,'' in \emph{Proceedings of the IEEE/CVF Conference on Computer
  Vision and Pattern Recognition (CVPR)}, 2019, pp. 9404--9413.

\bibitem{wu2019detectron2}
Y.~Wu, A.~Kirillov, F.~Massa, W.-Y. Lo, and R.~Girshick, ``Detectron2,''
  \url{https://github.com/facebookresearch/detectron2}, 2019.

\bibitem{Mikolov2013EfficientEO}
T.~Mikolov, K.~Chen, G.~Corrado, and J.~Dean, ``Efficient estimation of word
  representations in vector space,'' in \emph{Proceedings of the International
  Conference on Learning Representations (ICLR)}, 2013.

\bibitem{Newbury2021LearningTP}
R.~Newbury, K.~He, A.~Cosgun, and T.~Drummond, ``Learning to place objects onto
  flat surfaces in upright orientations,'' \emph{IEEE Robotics and Automation
  Letters (RA-L)}, vol.~6, pp. 4377--4384, 2021.

\end{thebibliography}

\end{document}